\documentclass{ecai} 
\usepackage{hyperref}
\usepackage[small]{caption}
\usepackage{latexsym}
\usepackage{amssymb}
\usepackage{amsmath}
\usepackage{amsthm}
\usepackage{booktabs}
\usepackage{enumitem}
\usepackage{graphicx}
\usepackage{color}
\usepackage{algorithm}
\usepackage{algorithmic}
\usepackage{tabularx}
\usepackage{multirow}

\usepackage{dashrule}
\usepackage{arydshln}
\usepackage{hyperref} 

\begin{document}

\begin{frontmatter}


\paperid{5924} 



\title{SDQ-LLM: Sigma-Delta Quantization for 1-bit LLMs of any size}

\author[A]{\fnms{Junhao}~\snm{Xia}\orcid{0009-0002-5586-7149}\thanks{Corresponding Author. Email:
xia-jh23@mails.tsinghua.edu.cn}}
\author[B]{\fnms{Ming}~\snm{Zhao}}
\author[B]{\fnms{Limin}~\snm{Xiao}} 
\author[B]{\fnms{Xiujun}~\snm{Zhang}}
\address[A]{Dept of Electronic Engineering, Tsinghua University, Beijing, China}
\address[B]{Beijing National Research Center for Information Science and Technology, Beijing, China}

\begin{abstract}
Large language models (LLMs) face significant computational and memory challenges, making extremely low-bit quantization crucial for their efficient deployment. In this work, we introduce SDQ-LLM: Sigma-Delta Quantization for 1-bit LLMs of any size, a novel framework that enables extremely low-bit quantization of LLMs while preserving their linguistic reasoning capabilities. A distinctive feature of SDQ-LLM is the continuous adjustability of the Over-Sampling Ratio (OSR), enabling dynamic adaptation to memory or VRAM constraints by selecting fractional OSR (e.g., 2.5×) for an optimal trade-off between model size and accuracy. SDQ-LLM uses upsampling combined with Sigma-Delta Quantizer to binarize or ternarize LLMs’ weights, encoding high-precision parameters into 1-bit or 1.58-bit representations, replacing the multiplication operations within linear layers with addition. This approach significantly enhances inference efficiency under extremely low-bit quantization. To further reduce the loss of quantization precision, we incorporate Hadamard-based weight smoothing prior to quantization, improving the stability and robustness of the weight representations. Furthermore, to fully leverage the continuity of the OSR and reduce precision loss, recognizing the correlation between quantization sensitivity and weight variance, we propose a fine-grained, layer- and linear-wise OSR allocation strategy, MultiOSR. This strategy distributes OSR both across layers and within each layer, based on weight variance and parameter scale. Finally, extensive experiments on OPT and LLaMA model families demonstrate that SDQ-LLM achieves a more efficient and high-precision performance even under highly aggressive low-OSR settings. Our code is available at \url{https://github.com/Dreamlittlecat/LLM-Quant-Factory}.

\end{abstract}

\end{frontmatter}


\section{Introduction}

Large language models (LLMs) based on transformers \citep{vaswani2017attention} have been a revolution in the AI field in recent years, with their surprising capabilities attracting increasing attention. Despite the impressive abilities of LLMs, their massive parameter scale and computational overhead make them difficult to deploy in resource-constrained environments. For example, the 70B LLaMA \citep{touvron2023llama} model requires at least around 150GB of memory for inference at half-precision(FP16). The huge memory footprint and resource consumption severely hinder the application of Large language models (LLMs) in edge-side scenarios.  

To tackle the storage and computational burden of large-scale parameters in LLMs, the LLM model compression field \citep{zhu2023survey} has witnessed rapid progress, with methods like parameter quantization, network pruning, low-rank decomposition, and knowledge distillation. Among these, LLM parameter quantization stands out as one of the most prevalent techniques.

Model quantization, a key deep learning optimization method, cuts model storage and computational complexity by converting parameters into low-bit formats, such as uniform/non-uniform quantization, binarization, and ternarization. Strategies like pretraining, optimization algorithms, and fine-tuning help preserve performance, mainly through Post-Training Quantization (PTQ) and Quantization-Aware Training (QAT).
QAT methods, e.g., LLM-QAT \citep{liu2024llmqat}, insert fake quantization nodes during training and use straight-through estimators to approximate gradients, enabling the model to adapt to low-bit representations early and reducing accuracy loss. However, due to extra training costs, PTQ remains more popular. PTQ converts a pretrained model's weights and activation values to low-bit formats post-training, using algorithms based on the model's statistical properties. 

At the same time, extremely low-bit model quantization, such as binarization and ternarization, represents special cases of the above quantization methods, where network parameters are restricted to very limited discrete values. In binarization, parameters are typically represented as binary values (e.g., -1 and +1) using the sign function. Representative methods include PB-LLM \citep{yuan2024pbllm}, BiLLM \citep{pmlr-v235-huang24q}, and OneBit \citep{DBLP:conf/nips/Xu0YWZLLC24}, which apply binary quantization to large language models to minimize memory and computation. In contrast, ternarization extends this approach by allowing an additional zero state (e.g., -1, 0, +1), offering a better trade-off between accuracy and efficiency. A representative ternarization-based approach is BitNet b1.58 \citep{ma2024era}, which achieves improved performance under extremely low-bit quantization settings.
Extremely low-bit quantization of model parameters is significant for hardware deployment, as techniques like binarizing or ternarizing parameters can convert multiplication operations in linear transformations to addition operations, greatly reducing hardware computation costs. However, extremely low-bit quantization often leads to considerable accuracy loss, making it difficult to apply in practice.

In the face of the formidable challenges posed by extremely low-bit quantization in large language models, we introduce the SDQ-LLM, Sigma-Delta Quantization for 1-bit LLMs of any size. By ingeniously repurposing the upsampling quantization principle rooted in sigma-delta analog-to-digital converters, SDQ-LLM pioneers a novel paradigm for parameter quantization in LLMs. Our research culminates in four key contributions that significantly advance the field: 
\begin{itemize}
    \item We propose SDQ-LLM, an upsampling-based Sigma-Delta quantization scheme designed for LLM parameters. This method enables extremely low-bit quantization while maintaining continuous control over both precision and model size. By adjusting the Over-Sampling Ratio (OSR), SDQ-LLM achieves a flexible balance between accuracy and compression, making it adaptable to a wide range of requirements and ensuring optimal performance across different use cases.
    \item To boost SDQ-LLM's quantization precision, we introduce the Hadamard matrix for weight smoothing. This technique reduces quantization errors and noise, ensuring more stable and reliable weight representations and improving the quantized model's overall quality. 
    \item We introduce MultiOSR, a weight-aware OSR allocation strategy that assigns OSR at both the layer and linear levels based on weight variance. This enables more efficient and targeted OSR distribution, enhancing the performance of the quantized model. 
    \item Through comprehensive experiments on OPT and LLaMA models, benchmarking against mainstream quantization techniques like RTN,GPTQ\citep{frantar2023optq}, PB-LLM\citep{yuan2024pbllm}, and BILLM\citep{pmlr-v235-huang24q}, and evaluating accuracy degradation, we prove SDQ-LLM's effectiveness and practicality in aggressive low-bit settings. 
\end{itemize}

\section{Related work}

In this section, we review related work from two aspects: model quantization and extremely low-bit quantization. First, we focus on mainstream Post-Training Quantization (PTQ) methods, renowned for their practicality and efficiency. Then, we introduce classic binarization and ternarization techniques, offering a detailed overview of these approaches. 

\subsection{ Model Quantization}

In model quantization, various effective methods aim to enable low-bit storage while maintaining accuracy. GPTQ \citep{frantar2023optq} reduces quantization errors by compensating unquantized parameter updates. AWQ \citep{MLSYS2024_42a452cb}, leveraging the varying importance of weights, uses an activation-aware approach to boost the scale of sensitive weights in RTN grouped quantization, minimizing errors. SpQR \citep{dettmers2024spqr} employs the Hessian matrix to separate outliers in weight distributions, preserving precision. OmniQuant \citep{shao2024omniquant} optimizes RTN quantization with learnable clipping coefficients and scaling factors. 
However, despite their success in conventional low-bit quantization under the RTN concept, these methods struggle with extremely low-bit cases like 1-bit quantization, where significant accuracy loss remains a major challenge. \\

\subsection{Extremely Low-bit Quantization}
Binarization and ternarization are typical extremely low-bit quantization methods. 
Binarization of parameters involves converting model parameters (weights and activation values) from a higher precision data type into only two discrete values (e.g., 1 and -1, or 0 and 1). Existing binarization methods, such as OneBit \citep{DBLP:conf/nips/Xu0YWZLLC24}, decompose parameter weights using SVD (Singular Value Decomposition), representing the original weights with high-precision row and column vectors and a binary sign matrix. To reduce quantization accuracy loss, OneBit \citep{DBLP:conf/nips/Xu0YWZLLC24} also adjusts the update vector and sign matrix using knowledge distillation. Ternarization of a model involves representing model weights with three discrete values (typically 1, 0, -1). 
As an updated version of BitNet \citep{wang2023bitnet}, BitNet b1.58 \citep{ma2024era} is among the few low-bit models adopting ternary quantization, where the Straight-Through Estimator (STE) is employed to enable effective training and to explore the model’s potential at extremely low bit-widths. Both OneBit\citep{DBLP:conf/nips/Xu0YWZLLC24} and BitNet b1.58 \citep{ma2024era} require additional training, making them QAT algorithms. In the PTQ domain (post-training quantization), binarization methods such as PB-LLM \citep{yuan2024pbllm} use a mask matrix to select and preserve the accuracy of highly sensitive weights, implementing mixed-precision quantization of the weights. BiLLM \citep{pmlr-v235-huang24q} uses the structural distribution of significant weights within the weight matrix (with significant weights primarily concentrated in a few columns) to implement a structured mask, reducing bit overhead for quantization and grouping nonsignificant weights into binary forms based on their bell-shaped distribution, improving quantization loss.

Nonetheless, these methods incur substantial accuracy degradation owing to the restricted precision in weight representation. To tackle this predicament, we put forward SDQ-LLM, which offers an appropriate trade-off between model size and accuracy. 
\section{Method}

This section details SDQ's quantization principles. First, it introduces the mathematical theories of RTN quantization, binarization, and ternarization. Then, it elaborates on the principles and process of Sigma-Delta Quantizer. Next, it explains the Hadamard matrix transformation for better accuracy. After that, the layer- and linear-wise OSR allocation strategy—MultiOSR is presented. Finally, the  entire Sigma-Delta Quantization algorithm is summarized.  

\begin{figure}[h]
\centering
\includegraphics[width=0.9\linewidth]{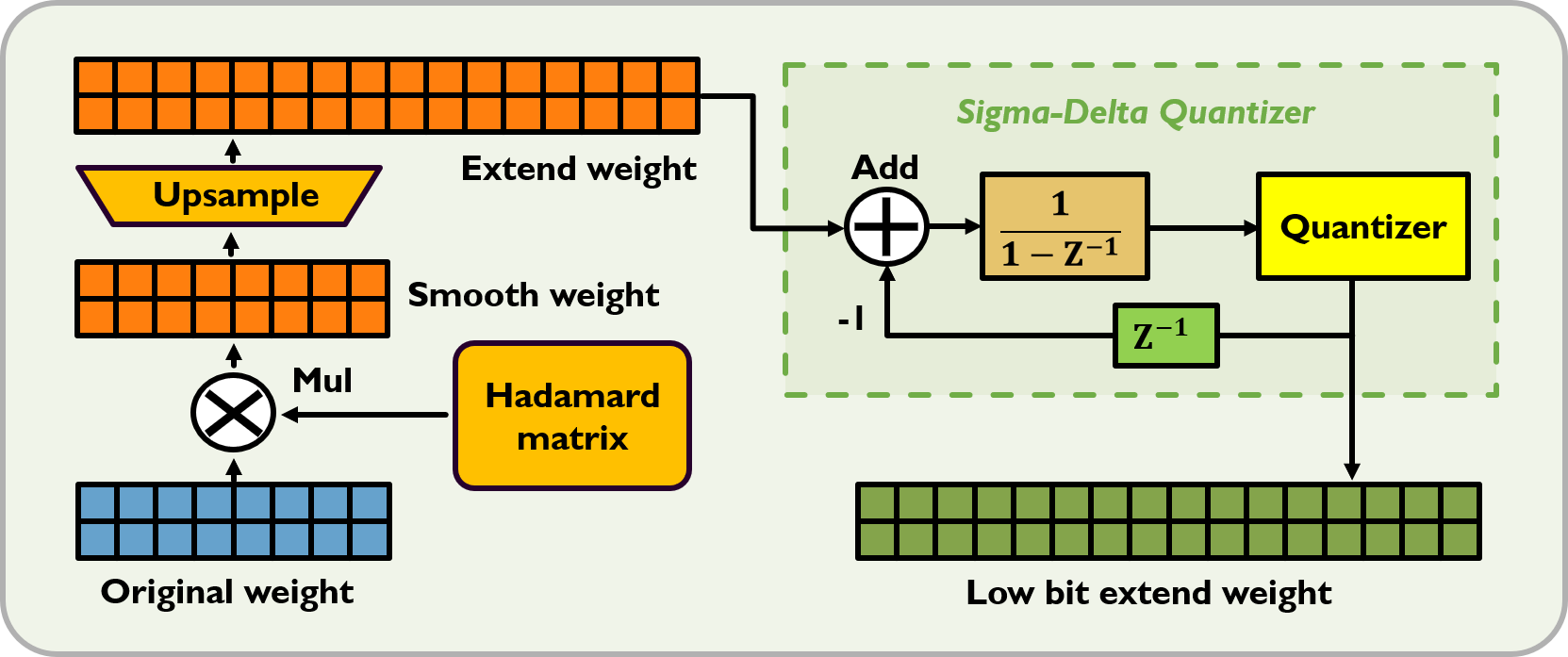}
\caption{Schematic Diagram of SDQ-LLM Processing Pipeline. The green dashed-line box represents the original Sigma-Delta Quantizer. $Z^{-1} $stands for a delay element and$\frac{1}{(1 - Z^{-1})}$  represents an integrator.  }
\label{fig:sigmadelta}
\end{figure}

\subsection{Preliminaries}\label{preliminaries}

Round-To-Nearest (RTN) quantization is the most fundamental and straightforward post-training quantization (PTQ) method. More advanced techniques such as GPTQ, AWQ, and OmniQuant are all built upon this basic principle. The core idea is to convert floating-point numbers to low-bit integers by rounding them to the nearest representable value. This is typically achieved through a linear transformation that maps floating-point values to integers, such as INT4 or INT8.

In practice, two quantization schemes are commonly employed: symmetric and asymmetric quantization. Symmetric quantization centers the mapping around zero, while asymmetric quantization does not require zero-centered values. However, the latter necessitates subtracting the minimum value from the original range in order to align the zero point.

Taking symmetric quantization as an example, the quantization process is governed by the following equations (1) and (2), where the round function denotes a standard rounding operation that maps a real number to its nearest integer:

\begin{align}
    W_{q} &=\Delta \cdot \text{Round} \left( \frac{W}{\Delta} \right)\\
    \Delta&=\frac{max(|W|)}{2^{N-1}}
\end{align}

Binarization and ternarization are model quantization methods that constrain the weights or activations of a neural network to a few discrete values, reducing computational complexity and storage. In binarization, values are limited to two levels, typically $\{-1, +1\}$, using the sign function:

\begin{align}
W_b &= \alpha \cdot \text{sign}(W),\\
\text{sign}(x) &=
\begin{cases}
+1 & x \ge 0,\\
-1 & x < 0,
\end{cases}
\end{align}

Ternarization extends this idea by introducing a third value, usually 0, to balance precision and efficiency, constraining weights or activations to $\{-1, 0, +1\}$:

\begin{align}
W\_t &= \alpha \cdot \text{Quant}(W),\\
\text{Quant}(x) &=
\begin{cases}
+1 & x > \theta,\\
0 & -\theta \le x \le \theta,\\
-1 & x < -\theta.
\end{cases}
\end{align}

In both cases, $\alpha$ is the quantization scaling factor, typically defined as $\mathrm{mean}(|W|)$, and $\beta$ is the quantization threshold, usually chosen as a fraction of $\alpha$, e.g., $\beta = 0.5\,\alpha$.

\subsection{Sigma-Delta Quantizer}\label{sdq}

\begin{figure}
    \centering
    \includegraphics[width=0.8\linewidth]{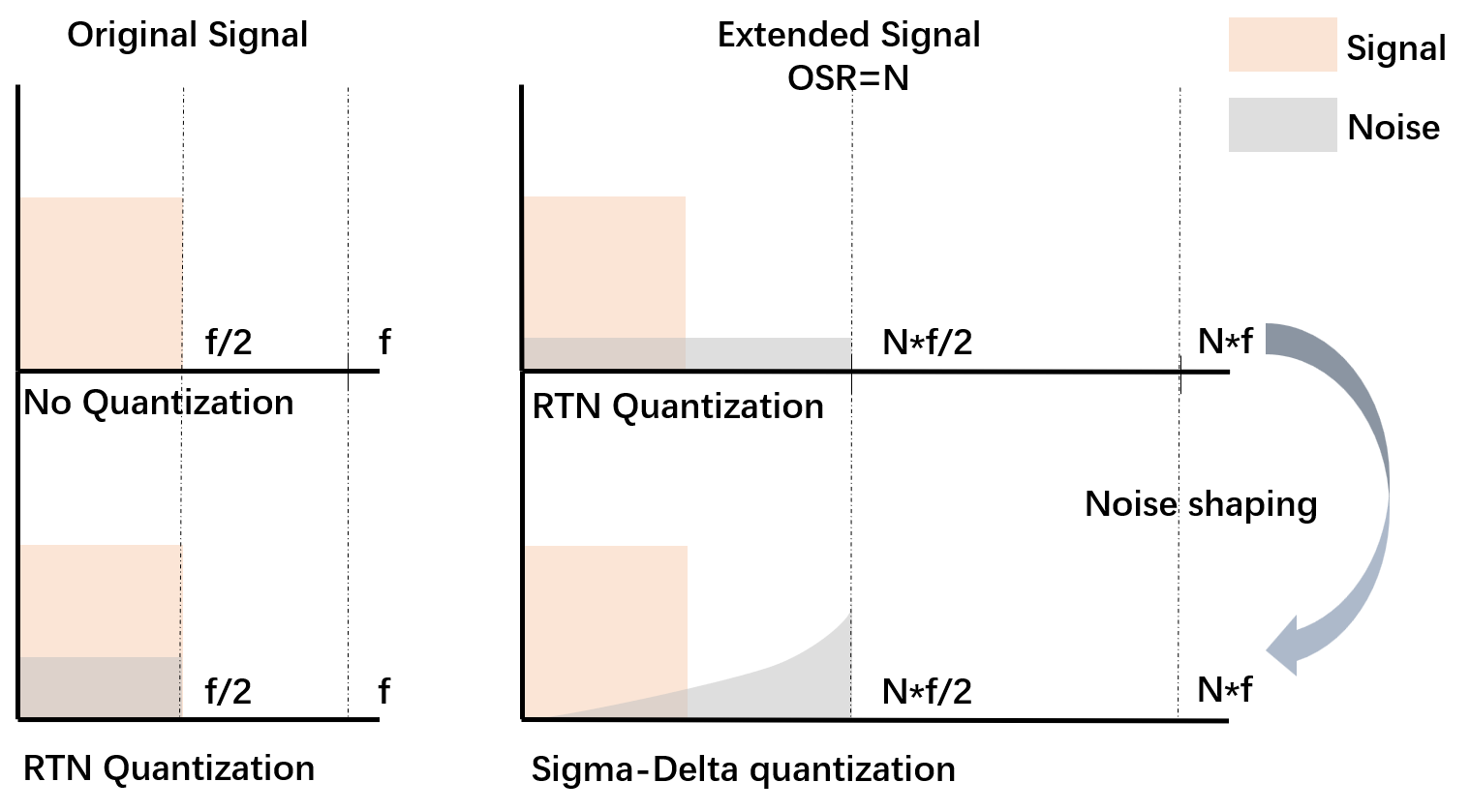}
    \caption{ Oversample and noise shaping. The left panel presents the spectrograms of the original signal before and after quantization, whereas the right panel presents those of the upsampled signal before and after quantization. }
    \label{fig:noise_shape}
\end{figure}

Sigma-Delta(\(\Sigma\text{-}\Delta\)) quantization is a technique used for converting analog signals into digital signals, widely applied in fields such as analog-to-digital converters (ADC) and digital audio processing. The core idea of this approach lies in enhancing quantization precision through the utilization of noise shaping and oversampling techniques. \textbf{Oversampling}: In analog-to-digital conversion (ADC), quantization errors introduce noise. After oversampling, the quantization noise is spread over a wider frequency range, reducing the quantization noise in the low-frequency portion of the target signal, thereby improving the signal-to-noise ratio (SNR). \textbf{Noise shaping} is one of the core mechanisms of Sigma-Delta (\(\Sigma\text{-}\Delta\)) Quantizer.  It utilizes a feedback loop to push the quantization noise energy from the low frequency range to the high frequency range, thereby improving the signal-to-noise ratio of low-frequency signals. For a more in-depth understanding of the specific implementation and related mechanisms, the detailed information is vividly illustrated in Figure \ref{fig:noise_shape}. 

Specifically, the first order \(\Sigma\text{-}\Delta\) Quantizer is defined by the iteration:

\begin{align}
    i_n &=i_{n-1}+x_n-y_{n-1}  \\
    y_n &=Q(i_{n}) =i_n+e_n 
\end{align}

$i_n$ is the output of the integrator, with an initial value of $i_0 = 0$. The function $Q(u)$ is defined in equations (1), (4), and (6). For a more intuitive understanding from the frequency-domain perspective, the first-order $\Sigma\text{-}\Delta$ quantizer can be analyzed using the Z-transform as follows:

\begin{align}
    I&=Iz^{-1}+X-Yz^{-1}\\
    Y&=I+E 
\end{align}

Therefore, we obtain the expression for the output $Y$ in terms of the input $X$ and the noise $E$:

\begin{align}
Y &= X + (1 - z^{-1}) E
\end{align}

In equations (9)–(11), $X$ is the Z-transform of the input, $I$ the integrator, and $E$ the quantization noise. The transfer functions are $H_x = 1$ for the input and $H_e = 1 - z^{-1}$ for the noise. Since $H_e$ is a high-pass filter, the quantizer shapes the noise toward higher frequencies, improving in-band SNR and reducing quantization error.

In further derivations, we need to briefly introduce Parseval's theorem. Parseval's Inner Product Identity describes the inner product relationship of a signal in both the time and frequency domains. It is a generalized form of Parseval's Theorem. The corresponding Parseval's inner product identity is given as follows:

\begin{align}
    \sum_{n} a(n) b^*(n) = \frac{1}{N} \sum_{k} A(k) B^*(k)
\end{align}

The signals \( a(n) \) and \( b(n) \) represent time-domain signals, while \( A(k) \) and \( B(k) \) are their corresponding Fourier transform representations. The symbol \( * \) denotes the complex conjugate, and \( N \) is the length of the signal. 

If we regard the activation input as \( a \) and the quantized weights as \( b \), then due to the equivalence of inner products in the time and frequency domains, during the linear operation \( A \times W^\top \), the quantized weights \( b \) effectively pass through a low-pass filter formed by \( A \), which suppresses the high-frequency noise introduced by Sigma-Delta quantization. 

\subsection{Hadamard Matrix}\label{hadamard}

Methods such as Quarot \citep{ashkboos2024quarot}, Quip \citep{chee2024quip}, SpQR \citep{dettmers2024spqr}, and PB-LLM \citep{yuan2024pbllm} show that weight matrices often contain a small number of  outliers. If not handled, these outliers can cause significant quantization errors. Inspired by Quarot, we use a Hadamard matrix to smooth the weights, reducing the impact of outliers and enabling more stable and accurate quantization.

\begin{figure}[h]
    \centering
    \includegraphics[width=0.89\linewidth]{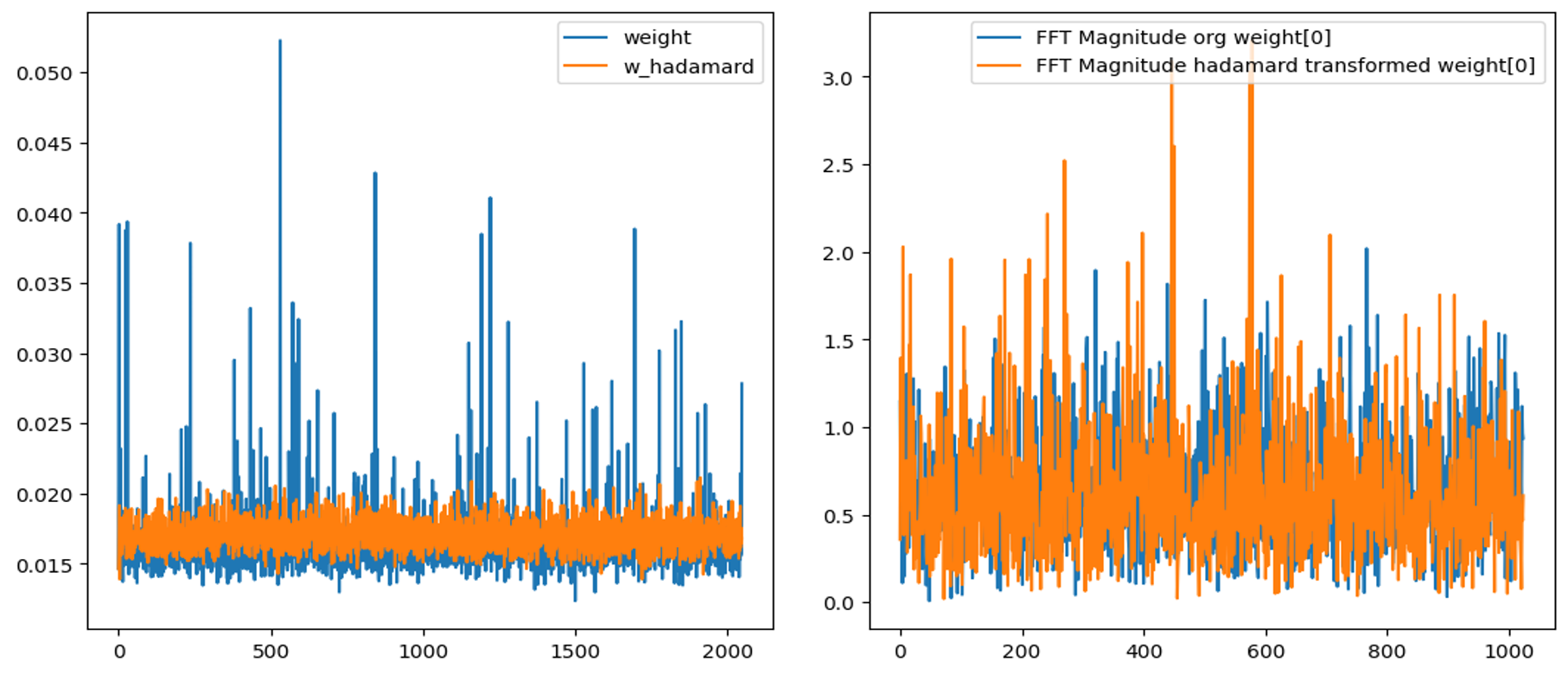}
    \caption{The time domain and frequency domain distribution of the opt-1.3b.layer.3.q\_proj weight matrix before (blue) and after (orange) being multiplied by the Hadamard matrix.}
    \label{fig:hadamard}
\end{figure}

A Hadamard matrix is a square matrix whose entries are either \( +1 \) or \( -1 \), and its rows (or columns) are mutually orthogonal. This means that the dot product of any two distinct rows (or columns) is zero. Formally, a matrix \( H \) is considered a Hadamard matrix if it satisfies the condition \( H H^T = nI \), where \( n \) is the size of the matrix and \( I \) is the identity matrix. Hadamard matrices can be constructed recursively, starting from a \( 1 \times 1 \) matrix, and expanding to larger matrices by using the following relation: 
\begin{align}
   H_{2k} = \begin{bmatrix} H_k & H_k \\ H_k & -H_k \end{bmatrix}
\end{align}


As shown in Fig \ref{fig:hadamard}, the left plot represents the distribution obtained by computing the column-wise mean of the absolute values of the weight matrix, while the right plot illustrates the frequency-domain distribution obtained by applying the FFT to the first row of the weight matrix. It can be observed that before multiplying by the Hadamard matrix, the weight matrix exhibits a certain proportion of outliers in the time-domain distribution. In the frequency domain, the distribution appears more uniform, with energy spread relatively evenly from low to high frequencies.

After applying the Hadamard transformation, the weight matrix becomes smoother in the time domain, which is reflected in the frequency domain as a concentration of energy in the low-to-mid-frequency range. This characteristic aligns remarkably well with Sigma-Delta quantization, which shapes quantization noise energy into the high-frequency range. This observation supports the rationale for incorporating the Hadamard transformation into the Sigma-Delta quantizer to enhance quantization accuracy.

\begin{algorithm}[t]
\caption{Quantize $\mathbf{W}$ given inverse hessian matrix $\mathbf{H}^{-1} = (2 \mathbf{X} 
\mathbf{X}^\top + \lambda \mathbf{I})^{-1}$ , block size $B$ , oversample ratio OSR and  Hadamard matrix Had  \(\in \mathbb{R}^{B \times B} \)}
\label{alg:sd-ptq}
\small

   \textbf{Input:} $\mathbf{W} \in \mathbb{R}^{d_\text{row} \times d_\text{col}}$ -- weight matrix \\
  \hspace{1.9em} $\mathbf{H}^{-1}$ -- inverse hessian matrix \\
 \hspace{1.9em} $B$ -- block size \\
 \hspace{1.9em} $OSR$ -- oversample ratio \\
 \hspace{1.9em} $Had$ -- random Hadamard matrix \\
 \textbf{Output:} $\mathbf{Q}$ -- quantized matrix 
\begin{algorithmic}[1]
    \STATE $\mathbf{Q} \gets \mathbf{0}_{d_\text{row} \times d_\text{col}}$ 
    \STATE $\mathbf{Q_{out}} \gets \mathbf{0}_{d_\text{row} \times (osr\cdot d_\text{col})}$
    \STATE $\mathbf{E} \gets \mathbf{0}_{d_\text{row} \times B}$ 
    \STATE $\mathbf{H}^{-1} \gets \text{Cholesky}(\mathbf{H}^{-1})^\top$ 
    \FOR {$i = 0, B, 2B, \dots$}
        \STATE $\mathbf{Q}_{:, i:i+B}, \mathbf{Q_{out}}_{:, i\cdot osr:(i+B)\cdot osr} \gets \text{SDQuant}(\mathbf{W}_{:, i:i+B} \cdot \text{Had},\text{OSR}) \cdot \text{Had}^\top$ 

        \STATE $\mathbf{E} \gets \mathbf{W}_{:, i:i+B} - \mathbf{Q}_{:, i:i+B}$ 
        \STATE $\mathbf{W}_{:, (i + B):} \gets \mathbf{W}_{:, (i + B):} - \mathbf{E} \cdot \mathbf{H}^{-1}_{i:(i + B), (i + B):}$ 
    \ENDFOR
    \STATE \textbf{return} $\mathbf{Q_{out}}$
\end{algorithmic}

\vspace{1em}
\hdashrule[0.5ex]{\linewidth}{0.5pt}{3pt 2pt}

\textbf{Function} \text{SDQuant}($\mathbf{W}$, $n$) \\
\textbf{Input:} $\mathbf{W} \in \mathbb{R}^{d_\text{row} \times d_\text{col}}$ -- weight block, $n$ -- oversampling ratio (OSR) \\
\textbf{Output:} $\mathbf{Y} \in \{-1,0, +1\}^{d_\text{row} \times (n d_\text{col})}$ -- binary quantized matrix

\begin{algorithmic}[1]
    \STATE $\mathbf{W} \gets \text{Resample}(\mathbf{W},\ n \cdot d_\text{col})$
    \STATE $\mathbf{Y} \gets \mathbf{0}$
    \STATE $\text{Integrator} \gets \mathbf{0}_{d_\text{row} \times 1}$
    \STATE $\text{Previous} \gets \mathbf{0}_{d_\text{row} \times 1}$
    \FOR{$i = 0, \dots, n d_\text{col} - 1$}
        \STATE $\text{Integrator} \gets \mathbf{W}_{:, i} - \text{Previous}$
        \STATE $\mathbf{Y}_{:, i} \gets \text{Quant}(\text{Integrator})$
        \STATE $\text{Previous} \gets \mathbf{Y}_{:, i}$
    \ENDFOR
    \STATE $\mathbf{Y_{out}} \gets \text{Resample}(\mathbf{Y}, d_\text{col})$
    \STATE \textbf{return} $\mathbf{Y_{out}}$,$\mathbf{Y}$
\end{algorithmic}

\vspace{1em}
\hdashrule[0.5ex]{\linewidth}{0.5pt}{3pt 2pt}
\textbf{Function} \text{Resample}($\mathbf{W}$, $N_\text{new}$) \\
\textbf{Input:} $\mathbf{W} \in \mathbb{R}^{d_\text{row} \times d_\text{col}}$ -- weight block, $N_\text{new}$ -- target column length \\
\textbf{Output:} $\mathbf{W}_\text{ext} \in \mathbb{R}^{d_\text{row} \times N_\text{new}}$ -- resampled weight block

\begin{algorithmic}[1]
    \FOR{$i = 0, \dots, d_\text{row}-1$}
        \STATE $\mathbf{w} \gets \mathbf{W}[i, :]$ \hfill 
        \STATE $X \gets \text{FFT}(\mathbf{w})$ \hfill 
        \STATE $Y \gets \text{ZeroPadOrTruncate}(X, N_\text{new})$ \hfill 
        \IF{$d_\text{col}$ is even}
            \STATE $Y[\frac{d_\text{col}}{2}] \gets
            \begin{cases}
                2 \cdot Y[\frac{d_\text{col}}{2}], & N_\text{new} < d_\text{col} \ (\text{downsample}) \\
                0.5 \cdot Y[\frac{d_\text{col}}{2}], & N_\text{new} > d_\text{col} \ (\text{upsample})
            \end{cases}$
        \ENDIF
        \STATE $\mathbf{w}_\text{ext} \gets \text{IFFT}(Y)$ \hfill 
        \STATE $\mathbf{w}_\text{ext} \gets \frac{N_\text{new}}{d_\text{col}} \cdot \mathbf{w}_\text{ext}$ \hfill 
        \STATE $\mathbf{W}_\text{ext}[i, :] \gets \mathbf{w}_\text{ext}$
    \ENDFOR
    \STATE \textbf{return} $\mathbf{W}_\text{ext}$
\end{algorithmic}

\end{algorithm}

\subsection{MultiOSR}\label{multiosr}

Recent studies like Layer-Wise Quantization \citep{dumitru2024layerwise} have demonstrated that different layers of large language models (LLMs) exhibit varying levels of sensitivity to quantization, and that assigning different precision across layers can effectively mitigate performance degradation. However, these efforts predominantly focus on layer-wise importance exploration, while systematic investigations at the finer-grained linear module level (e.g., within attention or feed-forward sublayers) remain scarce.

To further explore the continuity of OSR, we conduct a preliminary investigation into linear-wise OSR allocation. Our experimental findings indicate that the variance of model weights is roughly inversely related to quantization precision: smaller variance requires higher quantization precision. Here we present a possible explanation as follows: Low-variance weights, despite lower entropy, have higher information density and greater quantization sensitivity (small errors cause large noise), thus requiring higher OSR. 


Given the heterogeneous distribution characteristics across different linear modules within a layer, developing a distribution-aware, linear-wise precision (OSR) allocation strategies is crucial. Such approaches can more precisely capture the intrinsic sensitivity variations and achieve a better trade-off between compression efficiency and performance retention compared to coarse-grained layer-wise methods.

\begin{figure}[h]
\centering
\includegraphics[width=0.9\linewidth]{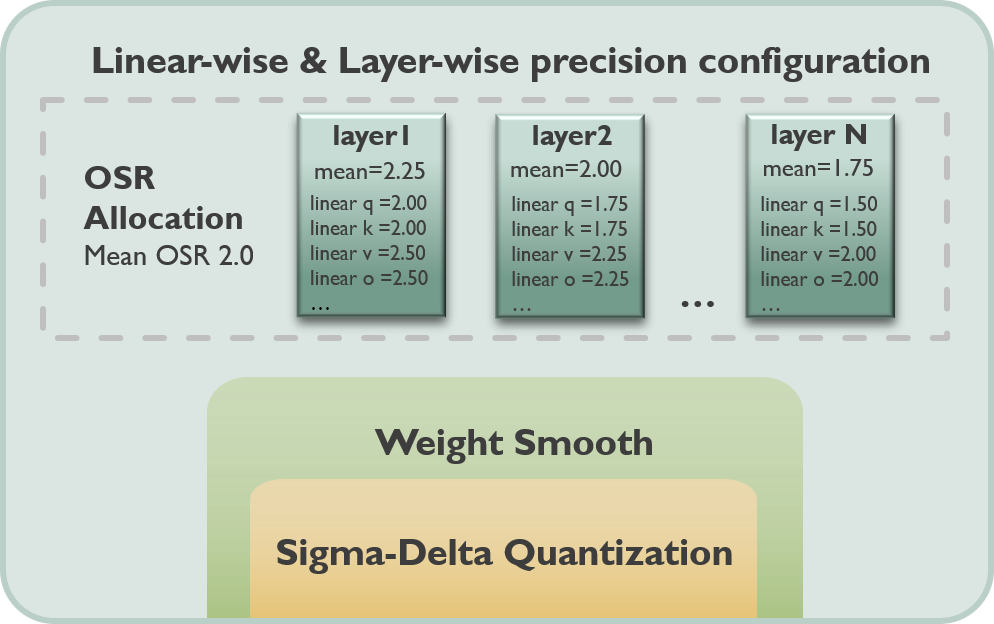}
\caption{The figure illustrates the MultiOSR allocation strategy: First, the average OSR for each decoder layer is computed based on the overall average OSR and the parameter variance of the respective layer. Then, within each decoder layer, the OSR for the linear layers (q, k, v, o, etc.) is assigned based on the layer’s average OSR, parameter variance, and weight proportion. }
\label{fig:multiosr}
\end{figure}

Motivated by the observed correlation between weight variance and quantization sensitivity, we propose MultiOSR, a layer- and linear-wise OSR allocation strategy (Figure \ref{fig:multiosr}), in which, given a target average OSR, layers are ranked by total weight variance, with OSR allocated accordingly. Within each layer, the allocated OSR is further distributed across linear modules based on two criteria: inversely proportional to weight variance and directly proportional to module size. By jointly considering both sensitivity and scale at the linear level, this fine-grained allocation achieves improved accuracy-efficiency trade-offs under aggressive quantization.

\subsection{Pipeline of SDQ-LLM}
\label{sec:Pipeline of SDQ-LLM}

\textbf{Quantization Workflow. }
As shown in Figure \ref{fig:multiosr}, the complete workflow of SDQ-LLM combined with MultiOSR is as follows: First, conduct an overall weight.-aware analysis of the LLMs to configure appropriate OSRs for different linear weights. Then, perform the quantization process shown in Figure \ref{fig:sigmadelta} to obtain the extended low-bit representation of the weights. In the actual algorithm implementation, by retaining only the block-wise compensation mechanism of GPTQ \citep{frantar2023optq} and eliminating column-wise quantization error compensation, we further reduce the quantization loss and improve the quantization accuracy. The complete quantization process is shown in Algorithm \ref{alg:sd-ptq}. In the algorithm, we choose a ternary function as the quantizer. 
Since the block compensation mechanism is introduced, the original output Y needs to be resampled to align with the original weight size.

\begin{figure}[h]
    \centering
    \includegraphics[width=0.69\linewidth]{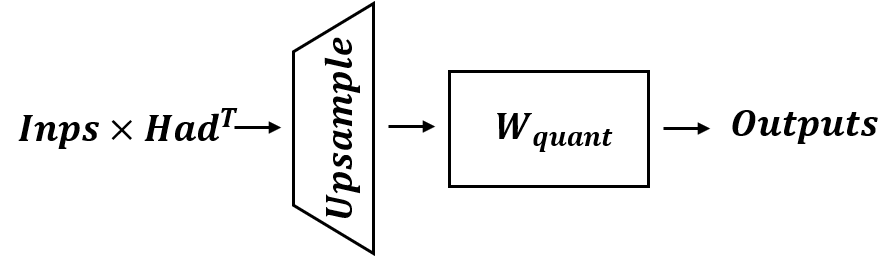}
    \caption{Flow chart of linear input during the inference process }
    \label{fig:x_upsample}
\end{figure}
 
As shown in Figure \ref{fig:sigmadelta}, since the size of the weights changes after quantization (Upsampling is applied, resulting in a size that is OSR times the original size. ), during actual inference, to ensure that the output is aligned with the size of the weights, the activations need to be upsampled with the same OSR value. The specific details are shown in Figure \ref{fig:x_upsample}. 

\begin{align}
    \eta &=\frac{N*OSR}{16} 
    \label{eq:compression ratio}
\end{align}

\textbf{Compression Ratio.}
The model storage cost of the SDQ-LLM method is controlled by the OSR and the selection of the specific quantizer. 
To quantify the storage efficiency of different settings, we introduce the notion of compression ratio. 
The compression ratio is defined with respect to the weights targeted for quantization, which typically include all linear modules within the transformer blocks, but exclude the embedding and language modeling (LM) head layers. It is computed as the ratio between the memory footprint of the quantized weights and that of the original full-precision weights. 
In the case of the SDQ method, the calculation of the compression ratio \(\eta\) of the weights follows the formula shown above, where \(N\) denotes the number of bits of the selected quantizer.

\section{Experiment}

This section presents experiments validating \textbf{SDQ-LLM}, covering setup, models, datasets, and comparisons with other low-bit methods. We include ablations on the Hadamard transform and MultiOSR, and show how continuous OSR controls the compression–performance trade-off.

\subsection{  Experimental Setup } 
All experiments are conducted in the PyTorch and CUDA environment, utilizing relevant tools from Hugging Face. The experiments are performed on a single NVIDIA RTX 4090 GPU. 

 \textbf{ Models and Datasets. } 
We evaluate on representative models from OPT \citep{zhang2022opt} (1.3B, 2.7B, 6.7B, 13B) and LLaMA \citep{touvron2023llama,touvron2023llama2,llama3} (LLaMA2-7B, 13B; LLaMA3.2-1B, 3B; LLaMA3-8B), covering a broad range of scales.
Regarding dataset selection, we use WikiText2 \citep{merity2016pointersentinelmixturemodels} and a subset of C4 \citep{JMLR:v21:20-074} for perplexity testing. WikiText2, drawn from Wikipedia, captures diverse language usage, while C4, a large-scale web corpus, is widely used for LM training. Following GPTQ, we adopt its compensation update and calibrate on a small subset of C4 with no test-set overlap.
In addition to PPL testing, we further evaluate SDQ-LLM on six zero-shot downstream tasks to examine its practical effectiveness across a range of real-world applications. The selected tasks include:
PIQA \citep{bisk2020piqa}
BoolQ \citep{clark2019boolqexploringsurprisingdifficulty}
Winogrande \citep{sakaguchi2021winogrande}
ARC-e  \citep{clark2018thinksolvedquestionanswering}
ARC-c \citep{clark2018thinksolvedquestionanswering}
and HellaSwag  \citep{zellers2019hellaswag}. 
These tasks span commonsense reasoning, question answering, and knowledge understanding. By combining PPL evaluation on standard datasets with zero-shot task performance, we can comprehensively verify the effectiveness and generalization of the SDQ-LLM across both intrinsic and downstream metrics. 

\textbf{Baseline Methods. } 
SDQ-LLM, an efficient post-training quantization (PTQ) framework, eliminates the need for fine-tuning and allows the entire process to be completed in a single quantization step. Given its efficiency, we primarily choose PTQ methods for comparative experiments. 
Specifically, we select vanilla RTN and GPTQ \citep{frantar2023optq} as references, and opt for relatively mainstream binarization methods like PB.-LLM \citep{yuan2024pbllm} and BILLM quantization \citep{pmlr-v235-huang24q} as comparison benchmarks.  

\renewcommand{\arraystretch}{1.5} 
\begin{table*}[t]
\caption{The table shows the perplexity (PPL \(\downarrow\)) of various quantization methods on WikiText2 across different model scales.  Models are grouped into OPT, LLaMA2, and LLaMA3, with superscripts $^{(2)}$ and $^{(3)}$ indicating LLaMA2 and LLaMA3, respectively. The weight bits of BiLLM are 1.11, 1.08, and 1.06 for the OPT, LLaMA2, and LLaMA3 series, respectively. 
}
\label{wikitext2}
\centering
\begin{tabular}{llllllllllll} 
\toprule
\multicolumn{3}{c}{\textbf{Setting}} &\multicolumn{4}{c}{\textbf{OPT}} & \multicolumn{5}{c}{\textbf{LLaMA2\&3}}  \\
\hline
\text{Method} & \shortstack{Block\\Size}  & \shortstack{Weight\\Bits} & 1.3B & 2.7B & 6.7B & 13B & ${7B}^{(2)}$ & ${13B}^{(2)}$ &  ${1B}^{(3.2)}$ &  ${3B}^{(3.2)}$ &  ${8B}^{(3)}$ \\
\hline
Full Precison  &- &16.00 & 14.62 &12.47 &10.86 &10.12 &5.47 &4.88 &9.75 &7.81 &6.13     \\

\hdashline
RTN  &- &2.00 &12782.84 &56577.08 &7831.01 &73564.32 &4460.33 &122.82 &151880.75  &118599.76 &284254.12 \\
GPTQ &128 &2.00 &107.65 &59.39 &21.03 &20.44 &63.22 &23.88 &5948.08  &5757.41 &890.19 \\
PB-LLM(10\%) &128 &1.70  &280.42 &144.37 &129.67 &85.17 &73.99 &139.77 &\textbf{270.11} &99.20 &69.57\\
BiLLM  &128 &\textbf{1.06\textasciitilde1.11} &70.06 &49.79 &47.24 &18.64 &29.06 &23.81 &1408.41 &152.36 &50.14\\

SDQ(OSR=2) &128 &1.58 &\textbf{38.24} &\textbf{20.30} &\textbf{14.87} &\textbf{11.60}   &\textbf{14.06} &\textbf{6.95}  &298.31 &\textbf{31.65} &\textbf{17.02}\\
\bottomrule
\end{tabular}
\end{table*}
\subsection{ Key Experimental Results. }
Our experiments aim to validate the feasibility and effectiveness of the proposed SDQ method. We evaluate the perplexity (PPL) performance of ternary quantization across RTN (group size 128), GPTQ (block size 128) \citep{frantar2023optq}, PB-LLM \citep{yuan2024pbllm}, BiLLM \citep{pmlr-v235-huang24q}, and SDQ, where the OSR for SDQ is set to 2. For PB-LLM, an outlier ratio of 10\% is used, with outliers selected via the Hessian matrix, and no QAT is applied. 
To further evaluate its practical utility, we test SDQ (OSR=2) on six zero-shot downstream tasks (PIQA \citep{bisk2020piqa}
BoolQ \citep{clark2019boolqexploringsurprisingdifficulty}
Winogrande \citep{sakaguchi2021winogrande}
ARC-e  \citep{clark2018thinksolvedquestionanswering}
ARC-c \citep{clark2018thinksolvedquestionanswering}
and HellaSwag  \citep{zellers2019hellaswag}) covering diverse reasoning and question-answering scenarios to assess its generalization beyond perplexity.

\textbf{Comparison Results. }
Table \ref{wikitext2} shows the PPL results on the WikiText2 dataset for various OPT and LLaMA models. When the OSR is set to 2 and a ternary quantization function is selected as the quantizer, SDQ consistently demonstrates superior performance in extremely low-bit quantization, outperforming traditional 1-bit and 2-bit quantization methods across all model sizes. The performance of RTN and GPTQ degrades significantly in comparison. As model size increases (e.g., LLaMA-13B and OPT-13B), SDQ shows a more pronounced accuracy advantage, demonstrating its effectiveness in preserving linguistic performance under extremely low-bit compression. 

\textbf{Zero-Shot Results. }
To comprehensively evaluate the practical utility of SDQ, we conduct experiments on six zero-shot downstream tasks: PIQA, BoolQ, Winogrande, ARC-e, ARC-c, and HellaSwag. These tasks encompass a wide spectrum of reasoning, commonsense understanding, and multiple-choice question-answering challenges. They serve as a rigorous and diverse testbed, enabling us to thoroughly assess the generalization ability of the model. Table \ref{zero_shot} demonstrate that SDQ exhibits highly competitive performance across these benchmarks. Even when subjected to highly compressed ternary quantization, SDQ holds its ground. It showcases that SDQ can effectively bridge the gap between theoretical evaluations and practical deployments, making it a promising approach for a variety of natural language processing tasks.

\renewcommand{\arraystretch}{1.5} 
\begin{table}[h]
\caption{Ablation results on Hadamard and MultiOSR are obtained using the LLaMA3‑8B model.}
\label{ablation}
\centering
\begin{tabular}{llllll} 
\toprule
\textbf{Method} &Hadamard  &MultiOSR    &WikiText2 \(\downarrow\) & C4 \(\downarrow\) \\

\hline
Full Precision &- &- &5.47 &7.26 \\
\hdashline
SDQ(OSR=2)  &N &N &2434.87 &708.39 \\
SDQ(OSR=2)  &Y &N &20.13 &26.26 \\
SDQ(OSR=2)  &N &Y &2751.13 &358.08 \\
SDQ(OSR=2)  &Y &Y &\textbf{17.02} &\textbf{24.72} \\
\bottomrule
\end{tabular}
\end{table}

\textbf{Ablation Results. }
In an endeavor to boost the efficacy of the SDQ method, we incorporate the Hadamard matrix to perform smoothing operations and adopt MultiOSR for a more optimized OSR configuration. To ascertain the effectiveness of these two components, we will conduct corresponding ablation experiments. As clearly illustrated in Table \ref{ablation}, which specifically focuses on the LLaMA3.-8B model, the performance of the SDQ method varies significantly depending on the different combinations of utilizing the Hadamard matrix and MultiOSR. According to the results, introducing the Hadamard matrix to smooth the weights has a very significant impact on the quantization accuracy. MultiOSR also improves the quantization accuracy to some extent, indicating that the research on MultiOSR has great potential. 

\renewcommand{\arraystretch}{1.5} 
\begin{table}[h]
\caption{Quantization Time (s) \(\downarrow\)}
\label{quantization time}
\centering
\begin{tabular}{llllll} 
\toprule
\textbf{Method} & OPT-1.3B &OPT-2.7B &OPT-6.7B  &OPT-13B \\
\hline
RTN  &10.02 &18.23 &37.46 &77.05 \\
\hdashline
GPTQ &90.16 &167.18 &337.95 &616.74 \\
PB-LLM(10\%)  &141.22 &243.98 &445.78 &778.49 \\
BiLLM  &361.49 &629.44 &1153.71 &1981.69\\
SDQ(OSR=2)  &\textbf{71.10} &\textbf{135.87} &\textbf{288.01} &\textbf{540.11}\\
\bottomrule
\end{tabular}
\end{table}

\textbf{Quantization Time. }
To more comprehensively verify the effectiveness of the SDQ method, we conducted a comparison of quantization times on the OPT series of models. Table \ref{quantization time} show that SDQ significantly outperforms other methods in terms of quantization time across all scale models. This demonstrates that SDQ has a distinct advantage in quantization efficiency, substantially reducing the quantization time and being more time-effective and efficient in practical applications.

\renewcommand{\arraystretch}{1.5} 
\begin{table*}[t!]
\caption{Zero-shot performance on Common Sense Reasoning tasks }
\label{zero_shot}
\centering
\begin{tabular}{llll llll lll}
\toprule
\textbf{Model} & \textbf{Method} & \textbf{\shortstack{Block\\Size}} &\textbf{\shortstack{Weight\\Bits}} & \textbf{ARC-c \(\uparrow\)} & \textbf{ARC-e \(\uparrow\)}  & \textbf{BoolQ \(\uparrow\)} & \textbf{Hellaswag \(\uparrow\)} &\textbf{PIQA \(\uparrow\)}  & \textbf{Winogrande \(\uparrow\)} &\textbf{Avg \(\uparrow\)}\\  

\midrule
\multirow{4}{*}{OPT-6.7B} 
& Full Precision &- &16.00 & 30.63 & 65.65 & 66.05 & 50.48  & 76.38 & 65.27  &59.07\\
\cdashline{2-11}
& GPTQ  &128 &2.00 &22.52 &53.99 &42.93 &41.66 &68.87 &56.90  &47.81\\
& PB-LLM (10\%) &128 &1.70 &19.62 &32.23 &61.25 &28.19 &57.29 &50.82  &41.56\\
& BiLLM &128 &\textbf{1.11} &17.40 &32.49 &60.76 &28.76 &58.75 &49.17 &41.22\\
& SDQ (OSR=2) &128 &1.58 &\textbf{27.90} &\textbf{56.94} &\textbf{64.37} &\textbf{42.77} &\textbf{71.21} &\textbf{60.29} &\textbf{53.91}  \\
\hline
\multirow{4}{*}{LLaMA2-7B} 
& Full Precision &- &16.00 &39.84 &69.23 &71.16 &56.68 &78.34 &67.16 &63.73\\
\cdashline{2-11}
& GPTQ   &128 &2.00 &20.73 &32.40 &54.09 &31.93 &58.10 &52.95 &41.70\\
& PB-LLM (10\%) &128 &1.70 &19.62 &26.51 &62.26 &26.85 &54.18 &49.64 &39.84\\
& BiLLM &128 &\textbf{1.08 }&22.44 &37.87 &62.11 &30.77 &61.20 &\textbf{54.14} &44.75\\
& SDQ (OSR=2) &128 &1.58 &\textbf{22.52} &\textbf{43.18} &\textbf{64.28} &\textbf{39.67} &\textbf{64.63} &51.61 &\textbf{47.64}\\
\hline
\multirow{4}{*}{LLaMA3-8B} 
& Full Precision  &- &16.00 &50.42 &80.09 &81.31 &60.17 &79.65 &72.53 &70.69\\
\cdashline{2-11}
& GPTQ  &128 &2.00 &19.45 &25.46 &42.56 &26.55 &51.19 &49.56 &35.79\\
& PB-LLM (10\%)  &128 &1.70 &18.00 &32.32 &48.34 &28.55 &56.58 &50.90 &39.13\\
& BiLLM  &128 &\textbf{1.06} &19.28 &38.93 &61.22 &30.88 &59.08 &52.40 &43.63\\
& SDQ (OSR=2)&128 &1.58  &\textbf{24.14} &\textbf{48.69} &\textbf{65.47} &\textbf{42.37} &\textbf{65.07} &\textbf{55.88} &\textbf{50.27}\\
\bottomrule
\end{tabular}
\end{table*}

\textbf{Continuous Precision. }Continuous precision transformation is one of the core achievements of this paper. By controlling the OSR, quantization for models of arbitrary size and precision can be achieved. 
To fairly compare the post-quantization storage cost of SDQ with other methods, we introduce the concept of compression ratio. The compression ratio, as formally defined in Section \ref{sec:Pipeline of SDQ-LLM}, refers to the size of the quantized linear weights divided by that of the original full-precision weights. 
For SDQ, the compression ratio \(\eta\) of the linear weights is computed as shown in Equation (\ref{eq:compression ratio}). For example, when the over-sampling ratio (OSR) is set to 1.5 and a ternary quantization function is used (i.e., \(N = 1.58\)), the resulting compression ratio is \(\eta = 1.58 \times 1.5 / 16 = 14.81\%\). 
Similarly, when OSR is set to 2, the compression ratio increases to \(\eta = 1.58 \times 2 / 16 = 19.75\%\).
For BiLLM and PB-LLM, since the corresponding mask matrix incurs an additional 1-bit storage cost, we set \(N = N_{\text{bits}} + 1\) when calculating the compression ratio, yielding \(\eta = N / 16\). The resulting compression ratios are 13.18\% and 16.86\%, respectively. In contrast, GPTQ and RTN use 2-bit quantization, resulting in a compression ratio of 12.5\%. As we use the same block size and group size across methods, the influence of the scale factor is omitted in the compression ratio comparison. 
With the quantizer set as ternary quantization, through experiments, we present the curve showing how the PPL (Perplexity) of OPT-6.7B on Wikitext2 changes with the OSR(which is inversely related to compression ratio).
As depicted in Figure \ref{fig: OPT-6.7B Wikitext2 ppl_osr}, the corresponding curve follows a concave function. We can select the optimal balance point between memory and precision based on the curve. For any given model, an accuracy-versus-OSR curve can be obtained. Based on the available memory budget, an appropriate OSR can then be selected to minimize accuracy degradation as much as possible.




\begin{figure}[h]
    \centering
    \includegraphics[width=0.89\linewidth]{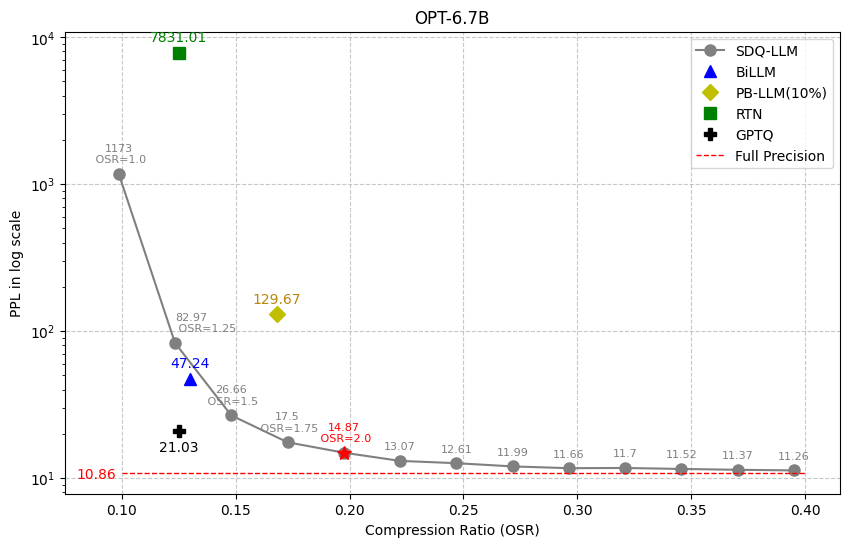}
    \caption{The  Wikitext2 PPL of the OPT-6.7B under different OSR. The OSR is incremented arithmetically from 1.0 to 4.0 with a step size of 0.25. The Sigma-Delta Quantizer employs a ternary quantization function. }
    \label{fig: OPT-6.7B Wikitext2 ppl&osr}
\end{figure}


\section{Conclusion}

In this work, we propose\textbf{ SDQ-LLM}, a novel arbitrary-precision extremely low-bit quantization method for LLMs. By introducing the concept of Sigma-Delta ADC, SDQ-LLM has achieved a model quantization method with continuously adjustable precision. To  reduce the quantization error introduced by quantization, we incorporate the \textbf{Hadamard transform} to smooth the weight matrix. This transformation redistributes the weight energy from a uniform frequency-domain distribution to a more concentrated low-to-mid-frequency distribution, effectively reducing quantization errors. To fully exploit the continuous variation of OSR, we propose \textbf{MultiOSR}, a layer- and linear-wise allocation strategy that leverages the correlation between weight variance and quantization sensitivity to systematically formulate and execute OSR assignments. 
\\



Ablation studies confirm the effectiveness of MultiOSR in reducing precision loss, while the results demonstrate that SDQ-LLM provides a robust solution for continuous quantization, maintaining considerable accuracy even under extremely low OSR settings and highlighting its potential in extremely low-bit quantization.

Considering the diverse application environments of LLMs, SDQ-LLM fits the compression ratio–accuracy curve to provide an optimal quantization strategy that balances memory usage and performance loss, further underscoring its potential for real-world deployment. By allowing flexible adjustment between compression and accuracy, it enables LLMs to be adapted to different hardware and application scenarios without retraining.

To the best of our knowledge, SDQ-LLM is the first quantization solution enabling continuous adjustment of model size and accuracy. Despite its advantages, it still faces significant challenges. In particular, there remains considerable room for exploration regarding the trade-off between size and precision at extremely low OSR, as well as the limits of performance retention under such aggressive compression.

\newpage









\bibliography{main}

\end{document}